\documentclass[letterpaper]{article}
\usepackage{aaai2026}
\usepackage{times}
\usepackage{helvet}
\usepackage{courier}
\usepackage[hyphens]{url}
\usepackage{graphicx}
\usepackage{natbib}
\usepackage{caption}
\usepackage{algorithm}
\usepackage{algorithmic}

\usepackage{newfloat}
\usepackage{listings}
\DeclareCaptionStyle{ruled}{labelfont=normalfont,labelsep=colon,strut=off}
\floatstyle{ruled}
\newfloat{listing}{tb}{lst}{}
\floatname{listing}{Listing}

\usepackage[utf8]{inputenc}
\usepackage[T1]{fontenc}
\usepackage{booktabs}
\usepackage{amsfonts}
\usepackage{nicefrac}
\usepackage{microtype}
\usepackage{xcolor}
\usepackage{colortbl}
\usepackage{afterpage}
\usepackage{subfigure}
\usepackage{subcaption}
\usepackage{multirow}
\usepackage{amsmath}
\usepackage{xspace}
\usepackage{pifont}

\def\ours{RelateSeg\xspace}

\definecolor{codegreen}{rgb}{0,0.6,0}
\definecolor{codegray}{rgb}{0.5,0.5,0.5}
\definecolor{codepurple}{rgb}{0.58,0,0.82}
\definecolor{backcolour}{rgb}{0.95,0.95,0.92}
\lstdefinestyle{mystyle}{
    backgroundcolor=\color{backcolour},   
    commentstyle=\color{codegreen},
    keywordstyle=\color{magenta},
    numberstyle=\tiny\color{codegray},
    stringstyle=\color{codepurple},
    basicstyle=\ttfamily\footnotesize,
    breakatwhitespace=false,         
    breaklines=true,                 
    captionpos=b,                    
    keepspaces=true,                 
    numbers=left,                    
    numbersep=5pt,                  
    showspaces=false,                
    showstringspaces=false,
    showtabs=false,                  
    tabsize=2,
    xleftmargin=15pt,                      
    xrightmargin=15pt,                     
    framextopmargin=10pt,                  
    framexbottommargin=10pt,               
}

\newcommand{\jy}[1]{{\color{black}#1}}

\newcommand{\gray}[1]{\textcolor{black}{#1}}

\title{
Neuro-Symbolic Spatial Reasoning in Segmentation
}

\author{
    Jiayi Lin\textsuperscript{\rm 1}, 
    Jiabo Huang\textsuperscript{\rm 2}, 
    Shaogang Gong\textsuperscript{\rm 1}
}
\affiliations{
    \textsuperscript{\rm 1}Queen Mary University of London\\
    \{jiayi.lin, s.gong\}@qmul.ac.uk \\
     \textsuperscript{\rm 2}Sony AI\\
     raymond.huang@sony.com \\
}

\begin{document}

\maketitle

\begin{abstract}
Open-Vocabulary Semantic Segmentation (OVSS) assigns pixel-level labels from an open set of categories, requiring generalization to unseen and unlabelled objects. Using vision-language models (VLMs) to correlate local image patches with potential unseen object categories suffers from a lack of understanding of spatial relations of objects in a scene. To solve this problem, we introduce neuro-symbolic (NeSy) spatial reasoning in OVSS. In contrast to contemporary VLM correlation-based approaches, we propose Relational Segmentor (RelateSeg) to impose explicit spatial relational constraints by first order logic (FOL) formulated in a neural network architecture. This is the first attempt to explore NeSy spatial reasoning in OVSS. Specifically, RelateSeg automatically extracts spatial relations, e.g., ⟨cat, to-right-of, person⟩, and encodes them as first-order logic formulas using our proposed pseudo categories. Each pixel learns to predict both a semantic category (e.g., "cat") and a spatial pseudo category (e.g., "right of person") simultaneously, enforcing relational constraints (e.g., a "cat" pixel must lie to the right of a "person"). Finally, these logic constraints are formulated in a deep network architecture by fuzzy logic relaxation, enabling end-to-end learning of spatial-relationally consistent segmentation. RelateSeg achieves state-of-the-art performance in terms of average mIoU across four benchmark datasets and particularly shows clear advantages on images containing multiple categories, with the cost of only introducing a single auxiliary loss function and no additional parameters, validating the effectiveness of NeSy spatial reasoning in OVSS.
\end{abstract}

\section{Introduction}

Open-Vocabulary Semantic Segmentation (OVSS) aims to segment an image into regions and assigns them labels from an open set of categories. 
Recent state-of-the-art methods~\cite{li2023open, diffseg, ovdiff} leverage pretrained vision-language models (VLMs)~\cite{clip, Stable_diffusion} to associate image regions with diverse textual concepts through data-driven correlations.
This reflects the strengths of modern deep learning, which excels at
pattern correlations but struggles with contextual reasoning and
structured understanding {--} traits characteristic of {System-1} in human
cognition~\cite{kahneman2011thinking, evans2013dual}.  
Humans, however, {also} engage in a more {contextually-relevant} 
{System-2 reasoning} that invokes rule-based reasoning  of structured knowledge in context.
To bridge this gap, neuro-symbolic (NeSy)
models~\cite{wang2024towards} {explores the spirit of combining}
the strengths of both {System-1 and System-2} by merging neural
perception ({System-1}) with symbolic reasoning
({System-2})~\cite{mcculloch1943logical}. {It aims} to unify
low-level data-driven visual recognition with high-level
knowledge-driven reasoning, mimicking the dual-recognition system of
human cognition.

\setlength{\fboxsep}{1pt}
\begin{figure}[t]
\centering
    \includegraphics[width=.45\textwidth]{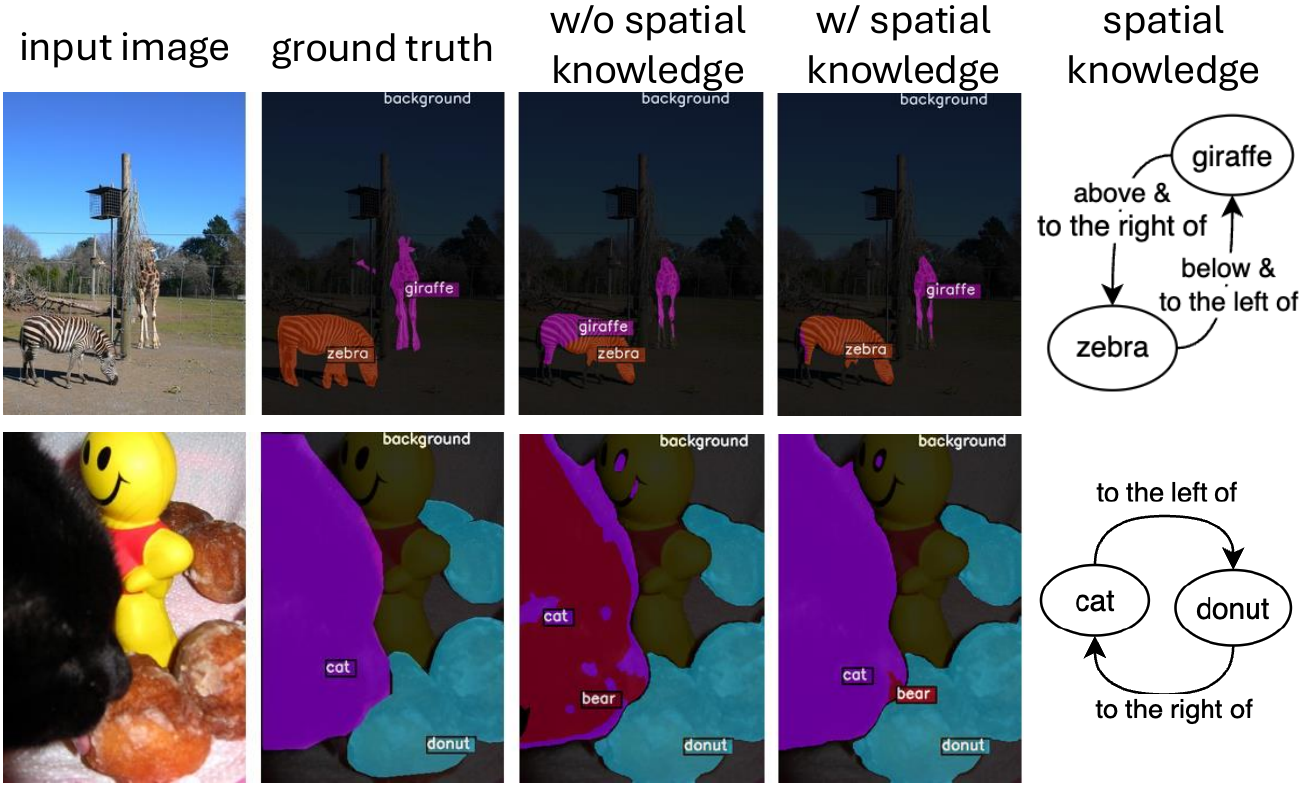} 
     \caption{Motivation illustration of \ours. The baseline model InvSeg~\cite{lin2025invseg} (third column), without incorporating spatial knowledge, can be confused between categories with similar features, such as zebra vs. giraffe and cat vs. bear, resulting in segmenting one object into two classes due to insufficient understanding of relative object positions. In contrast, \ours (fourth column) achieves clear separation of objects in different spatial positions by leveraging spatial knowledge.
  }

\end{figure}

Recently, several segmentation approaches~\cite{wang2019learning, bertinetto2020making, li2022deep} have incorporated neuro-symbolic (NeSy) frameworks, combining pre-defined symbolic knowledge of semantic concepts (e.g., complex meronymy and exclusion relations) as first-order logic (FOL) rules with neural networks.
However, these methods face two key limitations.
First, their reliance on manually pre-defined symbolic knowledge restricts applicability to closed-set scenarios, limiting scalability to diverse real-world applications.
Second, their symbolic constraints typically apply to individual pixels or objects in isolation, without modeling spatial scene structure—the relative positioning and contextual relations between objects (e.g., ``plates above tables"). 
Incorporating such spatial relations is crucial for minimizing implausible scene interpretation and producing logically consistent segmentation.

To address these limitations, 
we introduce \ours (Relational Segmentor), a NeSy segmentation model that represents spatial relations among objects in an image as  first-order logic formulas and incorporates them into network optimization.
Specifically,
we leverage vision-language models (VLMs) to reason about spatial relations {which are} represented as triplets in the form of $\langle \text{subject},
\text{relation}, \text{object} \rangle$. For example,
``A cat is to the right of a person" is represented as $\langle
\text{cat}, \text{right}, \text{person} \rangle$.
To incorporate logical constraints into network training,
contemporary NeSy methods \cite{wang2019learning, bertinetto2020making,
  li2022deep} use semantic hierarchy constraints that only apply to
individual pixels in isolation, 
which can then be
easily transformed into first-order logic (FOL) constraints
and later built into an objective loss function using fuzzy
logic relaxation~\cite{hajek2013metamathematics} for neural network optimization.
However, spatial relations involve
multiple pixels from different objects and positions, not individual pixels in isolation.
While spatial relation constraints can be formulated as FOL
constraints, they involve multiple pixels at different spatial
positions simultaneously, making fuzzy logic relaxation non-trivial
compared to only modelling single-pixel constraints.
To address this, we introduce pseudo categories to define
an object's adjacent areas of anchor objects. 
For example, a triplet $\langle \text{cat}, \text{right}, \text{person} \rangle$ can be expressed as:
$\forall x (\operatorname{Cat}(x) \to \operatorname{RightOfPerson}(x))$.
Since $\operatorname{RightOfPerson}(x)$ is derived from $\operatorname{Person}(x)$, the segmentation of category ``cat" depends on $\operatorname{Person}(x)$. 
This ensures the occurrence of ``cat" accompanied with
``person" on the right, enabling a logical constraint to be applied
across multiple pixels simultaneously. 

The contributions of this work {are:} 
\textbf{(1)}
For the first time, we explore NeSy spatial reasoning for open
vocabulary semantic segmentation. RelateSeg enables logical segmentation for diverse open-vocabulary scenes by automatically extracting spatial relations
using VLMs. 
\textbf{(2)} 
We introduce and construct spatial relation knowledge as symbolic constraints
for neuro-symbolic representation learning, solving the problem of
how to represent spatially distributed multiple pixels
simultaneously into logical constraints and trainable loss functions.
\textbf{(3)}
We demonstrate the benefits of NeSy spatial reasoning for OVSS by achieving state-of-the-art performance in terms of average mIoU across four benchmark datasets. In particular, our approach shows clear advantages on images containing multiple categories, with the cost of only introducing a single auxiliary loss function and no additional parameters.

\section{Related Work}

\noindent \textbf{Open Vocabulary Semantic Segmentation (OVSS):}
Early approaches relied on object-level cues such as boundaries and contours~\cite{malik2001contour, arbelaez2010contour}, while recent methods leverage web-scale image-text pretrained models~\cite{clip, Stable_diffusion}, enabling large-scale natural language supervision~\cite{maskclip, li2023open, wang2024image}.
These models surpass earlier ones based on attributes or word embeddings but depend heavily on image-text feature correlations, which are often unreliable at the pixel level due to limited structural reasoning.
We address this by integrating symbolic spatial knowledge from Vision-Language Models (VLMs), enabling more coherent segmentation through explicit reasoning about object positions and relations.

\noindent \textbf{Visual Spatial Relation:}
Understanding spatial relations is fundamental to scene interpretation, as established in foundational cognitive science works~\cite{marr1978representation, feldman2003visual}.
These studies suggest that object recognition involves not only physical attributes but also the spatial organization of objects, with humans integrating both bottom-up perceptual cues and top-down reasoning.
In computer vision,
spatial relations have benefited object
detection~\cite{chen2020improving} and action
recognition~\cite{wu2021spatial}, but remain underexplored for semantic segmentation.
Existing methods often handle single-object segmentation and rely on pre-segmented regions, limiting scalability~\cite{wu2024rgsan}.
A key challenge is representing spatial relations at the pixel level—while object detection can utilize spatial reasoning over region proposals~\cite{kim2021spatial}, enforcing such constraints pixel-wise is substantially more difficult.
We propose a neuro-symbolic learning model that incorporates spatial constraints directly into the loss function, refining masks without modifying network architecture, and enabling structured, relational understanding.

\begin{figure*}
  \includegraphics[width=0.99\textwidth]{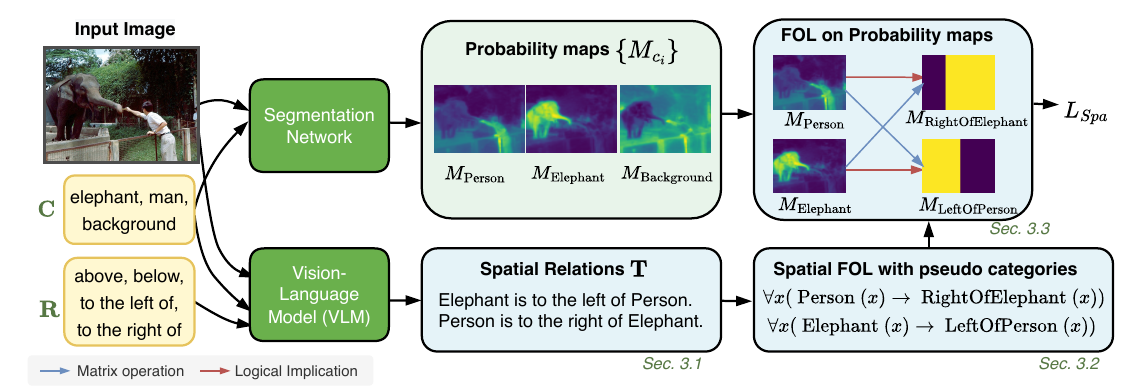}

\caption{
Overall framework of \ours.
Given a baseline segmentation model, e.g., InvSeg~\cite{lin2025invseg}, we apply spatial constraints
to the output probability maps $\{M_{c_i}\}$.
A VLM is used to extract a spatial triplet set $\mathbf{T}$ 
from the input image, based on a category set $\mathbf{C}$ and  a fundamental relation set $\mathbf{R}$.
These triplets are represented in First-Order Logic and incorporated into optimization, enabling a context-aware neuro-symbolic representation for segmentation.}
\end{figure*}

\noindent {\bf Neuro-Symbolic Visual Perception:}
Neural-symbolic (NeSy) systems ~\cite{mcculloch1943logical} combine the interpretability and reasoning capabilities of symbolic systems with the robust learning capacity of neural networks\cite{lake2017building, marcus2018deep}. 
Recent NeSy approaches in vision utilize category hierarchies as supervision and background knowledge~\cite{giunchiglia2021multi, giunchiglia2020coherent, wehrmann2018hierarchical}.
For example, HSS~\cite{li2022deep} models meronymy graph structures, while LogicSeg~\cite{li2023logicseg} encodes meronymy and exclusion constraints as differentiable rules.
However, these methods only explore symbolic rule at the individual pixel level and overlook structured reasoning.
Our method integrates bottom-up perceptual cues with top-down symbolic reasoning, enabling contextual object recognition that mirrors human perception, addressing limitations of purely connectionist models in capturing relational semantics.

\section{Methodology}
RelateSeg is a neuro-symbolic model for Open-Vocabulary Semantic Segmentation that incorporates spatial reasoning into segmentation. 
To this end,
\ours automatically extracts a set of spatial relations 
by iterative reasoning with a Visual Language Model (VLM).
%
After introducing pseudo category, these spatial relations are formalized as first-order logic rules. 
%
After a fuzzy-logic based continuous relaxation, 
such logic rules are formulated into a network architecture as a loss function, 
enabling logic-induced model training.
%

\subsection{Spatial Knowledge Acquisition}\label{sec:3.1}
Given a test image $I \in \mathbb{R}^{h \times w \times 3}$ and a list of categories from a test dataset,  
we first use a Visual Language Model (VLM)~\cite{li2022blip, liu2023visual} to extract a list of categories  
$\mathbf{C} = \{c_i \mid i = 1, 2, \ldots, C\}$ present in the image~\cite{diffseg, wang2025diffusion},  
including one category labeled as ``background''.
To acquire spatial relations, we consider every pair of categories $(c_i, c_j) \in \mathbf{C}$  
and use the VLM to infer their spatial relations.
We define a set of fundamental positional relations:  
$\mathbf{R} = \{\text{``above''}, \text{``below''}, \text{``left''}, \text{``right''}\}$
between pairwise categories
so that
more complex relations, such as ``top-right,'' are implicitly represented through combinations of these predefined relations  
(e.g., ``above'' + ``right'').

\noindent\textbf{Spatial relations inference.}
We first pass the image to the VLM and obtain the caption with prompt ``Describe this image in one sentence.''.
Then a question sequence is constructed from $c_i,c_j$ and a pair of opposite relation $r_k, r_l \in \mathbf{R}$.
For example, given $c_i=\text{``cat"}$, $c_j=\text{``person"}$, $r_k=\text{``left"}$ and $r_l=\text{``right"}$, the question is 
``[Caption] From the perspective of someone looking at the image, is the \textit{cat} positioned to the \textit{right}  or \textit{left} of the \textit{person}?''
We then extract all the relation in the answer.
In this way, we traverse through all $r_k, r_l$ for $c_i,c_j$.
Finally, we obtain a set of object pairs and the spatial relation between each pair, represented in a triplet  $\langle \text{subject}, \text{relation}, \text{object} \rangle$.
For example, the information ``The cat is to the right of the person" obtained from VLM can be represented in a triplet $\langle \text{cat}, \text{right}, \text{person} \rangle$.

\noindent\textbf{Spatial relations calibration.}
For each triplet, we create a bidirectional pair to complement the triplet set.
For example, given $\langle \text{cat}, \text{right}, \text{person} \rangle$, we add $\langle \text{person}, \text{left}, \text{cat} \rangle$ to the set.
Since VLM's answers are not always reliable, we implement several mechanisms to mitigate the impact of VLM performance limitations:  (1) Polar question validation:
We use polar questions,
where the prompt is
``[Caption] From the perspective of someone looking at the image, is the \textit{cat} positioned to the \textit{right} of the \textit{person}?''.
We only keep the triplets whose answer contain ``Yes''.
(2) Contradiction filtering:
We remove contradictory triplets to yield the final spatial knowledge. We define two types of contradictions:
(a) Cyclic contradiction: e.g.,
$\langle \text{person}, \text{right}, \text{cat} \rangle$ and $\langle \text{cat}, \text{right}, \text{person} \rangle$
(b) Directional contradiction: e.g.,
$\langle \text{person}, \text{right}, \text{cat} \rangle$ and $\langle \text{person}, \text{left}, \text{cat} \rangle$.
We collect all the contradictory pairs and ask VLM to select one using the same form of the first question.
Finally, we obtain the final spatial triplet set $\mathbf{T}=\{\langle s_t, r_t, o_t \rangle \mid t=1,2,\ldots,T; s_t, o_t \in \mathbf{C}\}$.

\subsection{Pseudo Category Based Knowledge Representation}\label{sec:3.2}


\noindent\textbf{Symbolic knowledge representation.}
Following previous work~\cite{li2022deep, li2023logicseg}, we use propositional logic to describe whether a pixel in an image belongs to a category.
For example,
the atomic proposition $\operatorname{Cat}(x)$ represents the statement “$x$ is a cat”, 
where the variable $x$ is a pixel in an image and the unary predicate $Cat$ represents the property of “being a cat”.
Further, the universal
quantifier $\forall$ asserts that a predicate holds for all pixels in
an image.
The first-order logic (FOL) proposition $\forall x (\operatorname{Cat}(x))$ 
means that for all pixels $x$, $x$ belongs to the category ``cat''.
Previous NeSy methods~\cite{wang2019learning, bertinetto2020making}
mainly use FOL
to apply semantic hierarchy constraints.
For example, the FOL $\forall x \, (\operatorname{Cat}(x) \to \operatorname{Animal}(x)) $,
 constraints a single pixel predicted
as ``cat'' should also be predicted as ``animal'' at the same time.
%
However, a spatial relation constraint naturally needs to involve pixels from different spatial positions at the same time. 
For example, the triplet $\langle \text{cat}, \text{right}, \text{person} \rangle$
 indicating that a cat is to the right of a person,
can be represented as:
\begin{equation}
\label{equ1}
\begin{alignedat}{2}
&\forall x \forall y \, (\operatorname{Person}(x) \land
\operatorname{Cat}(y) \to \operatorname{RightOf}(x,y)),
\end{alignedat}
\end{equation}
where the predicate 
$\operatorname{RightOf(x,y)}$
represents
``$x$ is to the right of $y$''.

\noindent\textbf{Spatial FOL with pseudo category.} 
To connect different spatial pixels in a FOL proposition,
we introduce a pseudo predicate/category 
$\{r\_of\_c \mid r \in \mathbf{R}, c \in \mathbf{C}\}$
to represent an object's adjacent area.
For example, $\operatorname{RightOfPerson}$ represents the right side area of a person.
Therefore, the triplet $\langle cat, right, person \rangle \in \mathbf{T}$ implies that any cat instance in the image must be located in the area to the right of the person. 
Formally, any pixel belongs to category ``cat'' must belong to the pseudo category $\operatorname{RightOfPerson}$, 
which defines the spatial region to the right of the person instance.
In this way, the triplet $\langle cat, right, person \rangle$
can be stated as:
\begin{equation}
\label{equ2}
\begin{alignedat}{2}
&\forall x \forall y \, (\operatorname{Person}(x) \land
\operatorname{Cat}(y) \to \operatorname{RightOfPerson}(y)).
\end{alignedat}
\end{equation}
Therefore, we ensure the appearance of ``cat'' should be accompanied with the appearance of ``person'' on the right at the same time, 
enabling the logical constraint to be applied 
on multiple pixels at the same time.
In practice, we derive the output segmentation mask of category $\operatorname{RightOfPerson}$ from the mask of $\operatorname{Person}$,
so $\forall x,y \, ( \operatorname{RightOfPerson}(y) \to
\operatorname{Person}(x) )$. Eq.~(\ref{equ2}) can then be
simplified as:  
\begin{equation}
\label{equ3}
\begin{alignedat}{2}
&\forall x \, 
(\operatorname{Cat}(x) \to \operatorname{RightOfPerson}(x)).
\end{alignedat}
\end{equation}

\subsection{FOL Incorporation for Network Optimization}\label{sec:3.3} 
We apply the FOL constraints like Eq.~(\ref{equ3}) to the output of a segmentation network,
which produces probability maps $\{M_{c_i}, i=1,2,...,C\}$ corresponding to the category list $\mathbf{C}$, where $M_{c_i}(x) \in [0,1]$ represents the probability of pixel $x$ belonging to category $c_i$. 

\noindent\textbf{Probability map for pseudo category.}
At the first place, we obtain the probability map of pseudo categories (like $\operatorname{RightOfPerson}$), which are derived from their original categories (like $\operatorname{Person}$).
For each original category, we can derive the four pseudo categories corresponding to the predefined positional relation set $\mathbf{R}$.
We take generating $M_{\operatorname{RightOfPerson}}(x)$ from $M_{\operatorname{Person}}(x)$ as an example.
Let $X \in \mathbb{R}^{H \times W}$ be the x-coordinate map that represent the x-coordinate of the current position.
For any position $(i,j)$ in the map:
$X_{i,j} = j$
where $i \in {1,\dots,H}$ and $j \in {1,\dots,W}$.
We then weight the x-coordinate map $X$ with $M_{\operatorname{Person}}(x)$ and sum up the output to get the mean x-coordinate of the $M_{\operatorname{Person}}(x)$:
\begin{equation}
\bar{x} = \frac{\sum_{i,j} X_{i,j} \cdot M_{\operatorname{Person}}(x_{i,j})}{\sum_{i,j} M_{\operatorname{Person}}(x_{i,j}) + \epsilon},
\end{equation}
where $\epsilon$ is a small positive constant added to prevent division by zero.
Finally, we define the right-of-person map  by comparing with this mean x-coordinate:
\begin{equation}
M_{\operatorname{RightOfPerson}}(x_{ij}) = \begin{cases}
1 & \text{if } X_{i,j} \geq \bar{x} \\
0 & \text{otherwise}
\end{cases}
\end{equation}
This creates a binary mask where all pixels with x-coordinates greater than or equal to the mean x-coordinate of the $M_{\operatorname{Person}}(x)$ are set to 1, and all others are set to 0.
We use the same strategy to generate probability maps for other three pseudo maps.

\noindent\textbf{Spatial loss function $\mathcal{L}_{Spa}$.}
Following previous work~\cite{li2023logicseg,sueyoshi2024predicated}, we employ fuzzy logic relaxation~\cite{hajek2013metamathematics, prokopowicz2017theory, van2022analyzing} to transform logical constraints into differentiable loss functions for network optimization. 
Fuzzy logic extends classical Boolean logic by allowing truth values to vary continuously between 0 and 1, 
making it suitable for handling approximate reasoning and uncertainty in neural network outputs.
Therefore, the probability maps $M_{c_i}(x)$ can be interpreted as continuous relaxations of logical propositions $\operatorname{c_i}(x)$.

We employ the product fuzzy logic and its operations.
Given fuzzy membership values $ \operatorname{P}(x), \operatorname{Q}(x) \in [0,1] $, which represent the degree to which a proposition is satisfied (0 for completely false, 1 for completely true).
We use the implication operation $\operatorname{P}(x) \to \operatorname{Q}(x) = 1 - \operatorname{P}(x) \times (1 - \operatorname{Q}(x))$.
The implication operation captures the intuition that a constraint is violated only when the premise $\operatorname{P}(x)$ is strongly satisfied while the conclusion $\operatorname{Q}(x)$ is not. For instance, when $\operatorname{P}(x)= 1 $ (premise fully true) and $\operatorname{Q}(x) = 0 $ (conclusion fully false), the implication equals 0 , indicating maximal violation; conversely, when  $\operatorname{P}(x)= 0 $ or  $\operatorname{Q}(x)= 1 $, the implication equals 1, indicating full satisfaction. 

We further use the universal quantifier $\forall$ approximation
$\forall x. P(x) = \prod_{k} A_P[k]$.
The universal quantifier is approximated as a product across spatial locations, ensuring the constraint is satisfied on average across pixel pairs rather than universally for every pair. This continuous, relaxed formulation allows us to incorporate approximate spatial relations as differentiable loss terms, making it suitable for real images where strict logical constraints may not hold everywhere.
Therefore, the loss function for Eq.~\ref{equ3} is: 

\begin{equation}
\begin{aligned}
\label{equ4}
\mathcal{L}[\forall x . & \operatorname{Cat}(x) \to \operatorname{RightOfPerson}(x)] \\
&=-\sum_{x_{i,j}} \log (1-M_{\operatorname{Cat}}(x_{i,j}) \times \\ 
&\qquad \qquad \qquad \qquad \left(1-M_{\operatorname {RightOfPerson}}(x_{i,j})\right))).
\end{aligned}
\end{equation}
We use the negative logarithm, inspired by the negative log-likelihood for Bernoulli random variables as in previous work~\cite{sueyoshi2024predicated}.
We denote the loss function for each FOL in the image as $l_1, l_2,\ldots,l_T$.

\noindent\textbf{Constraint-specific re-weighting.} We weight each loss function by the probability map score of the original category.
For example, for $\mathcal{L}[\forall x .  \operatorname{Cat}(x) \to \operatorname{RightOfPerson}(x)]$, its corresponding weight is:
\begin{equation}
\bar{x} = \frac{\sum_{i,j} M_{\operatorname{Person}}(x)(x_{ij}) \cdot \sigma(M_{\operatorname{Person}}(x))(x_{ij})}{\sum_{i,j} \sigma(M_{\operatorname{Person}}(x))(x_{ij}) + \epsilon}
\end{equation}
$\sigma$ is a sigmoid function with hyper-parameter: $bias=0.7, scale=10$.
We compute $\mathcal{L}_{Spa}$ obtaining the weighted sum of among $l_1, l_2,\ldots,l_T$.

\noindent\textbf{Overall optimization process.}
We build upon the current state-of-the-art method InvSeg~\cite{lin2025invseg} as our baseline. 
The spatial constraint loss $\mathcal{L}_{Spa}$ is added into InvSeg's original loss function with the weight $\alpha=0.1$ to optimize the model.


\begin{table*}[h!]

\centering
\resizebox{\textwidth}{!}{
\setlength{\tabcolsep}{12pt}
  \begin{tabular}{lcccccc}
    \toprule
    {\multirow{2}{*}{\centering{Methods}}} & 
    {\multirow{2}{*}{\centering{Venue}}} & 
    {\centering{PASCAL }} & 
      PASCAL  & 
      COCO  & \multicolumn{1}{c}{\multirow{2}{*}{\centering{ADE20K}}} &
      {\multirow{2}{*}{\centering{Average}}}
      \\ 
     &
     &
      {\centering{VOC}} & 
       Context & 
      Object & & \\
    \midrule
    \midrule
    
    {\color{gray}\emph{Training-based}}& & & & & & \\
    {\color{gray}TCL~\cite{tcl}}                       & {\color{gray}CVPR 2023}     & {\color{gray}51.2} & {\color{gray}24.3} & {\color{gray}30.4} & {\color{gray}17.1} & {\color{gray}30.8}  \\
    {\color{gray}OVSegmentor~\cite{ovsegmenter}}    & {\color{gray}CVPR 2023}   & {\color{gray}53.8} & {\color{gray}20.4} & {\color{gray}25.1} & {\color{gray}\, 5.6} & {\color{gray}26.2} \\
    {\color{gray}CoDe~\cite{wu2024image}}         & {\color{gray}CVPR 2024}     & {\color{gray}57.7} & {\color{gray}30.5} & {\color{gray}32.3} & {\color{gray}17.7} & {\color{gray}34.5} \\
    {\color{gray}CLIP-DINOiser~\cite{wysoczanska2024clip}}  & {\color{gray}ECCV 2024} & {\color{gray}62.2} & {\color{gray}32.4}  & {\color{gray}35.0} & {\color{gray}{20.0}} & {\color{gray}{37.4}} \\
    \midrule
    \emph{Training-free} & & & & & &\\
    CLIP~\cite{clip}                                & ICML 2021 & 16.4  & \, 8.4   & \, 5.6 &  \, 2.9 & \, 8.3 \\
    Stable Diffusion $\ast$~\cite{Stable_diffusion} &CVPR 2022 &  55.8&24.3&30.8&10.9 & 30.4 \\
    MaskCLIP~\cite{maskclip}                        & ECCV 2022 & 38.8 & 23.6 & 20.6 & \, 9.8 & 23.2 \\
    SCLIP~\cite{wang2024sclip} & ECCV 2024 & 59.1 & \textbf{30.4} & 30.5 & \textbf{16.1} & \underline{34.0 }\\
    CLIPSurgery~\cite{li2025closer} & PR 2025 & -&  29.3 & -  &  - & - \\
    DiffSegmenter~\cite{wang2025diffusion}   & TIP 2025 & {{60.1}}  & {{27.5}} & \textbf{37.9} & - & - \\
    \midrule
    \emph{Test time optimization} & & & & & &\\
    
Diffusion+TPT$\ast$~\cite{shu2022test} & NeurIPS 2022 & 60.3 & 28.4 & 33.4 & 12.1 & 33.5 \\
    InvSeg~\cite{lin2025invseg}    & AAAI 2025 & \textbf{63.4}& {27.8} & \underline{36.0} &- & - \\
    
\rowcolor{gray!10}\ours (Ours)              & - & \underline{63.1} &\underline{30.1} & 35.3 & \underline{13.0} & \textbf{{35.4}} \\
    \bottomrule
  \end{tabular}
  }
\caption{\textbf{Performance comparison on OVSS.} \textbf{Bold fonts} depict the best scores
and \underline{underline fonts} refer to second best. Notation: 
``--'': results unavailable. 
$\ast$: our implementation.}\label{tab:sota}
\end{table*}

\setlength{\fboxsep}{1pt}
\begin{figure*}[t!]
\centering
    \begin{subfigure}
        \centering
        \includegraphics[width=0.42\textwidth]{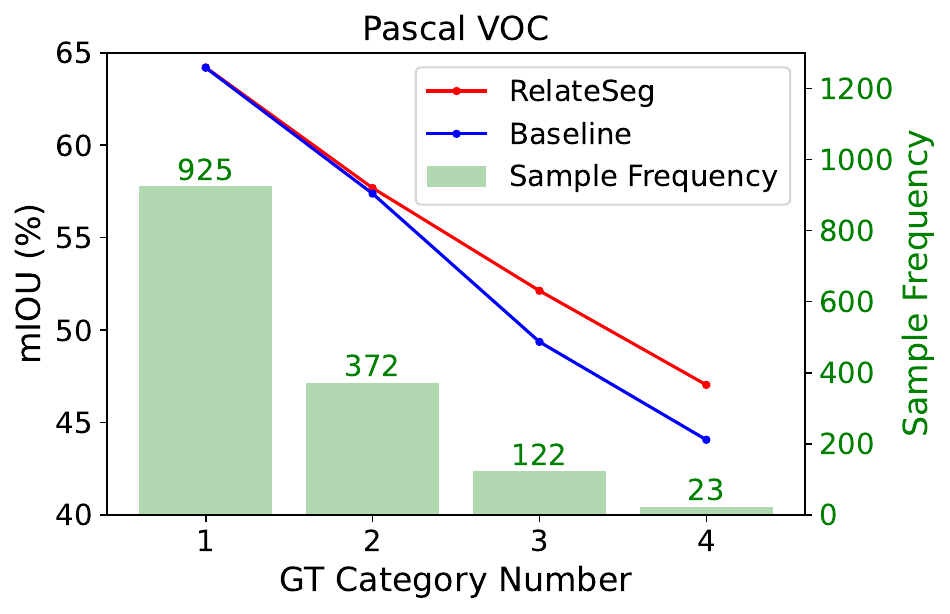}
    \end{subfigure}
    \hspace{10pt}
    \begin{subfigure}
        \centering
        \includegraphics[width=0.533\textwidth]{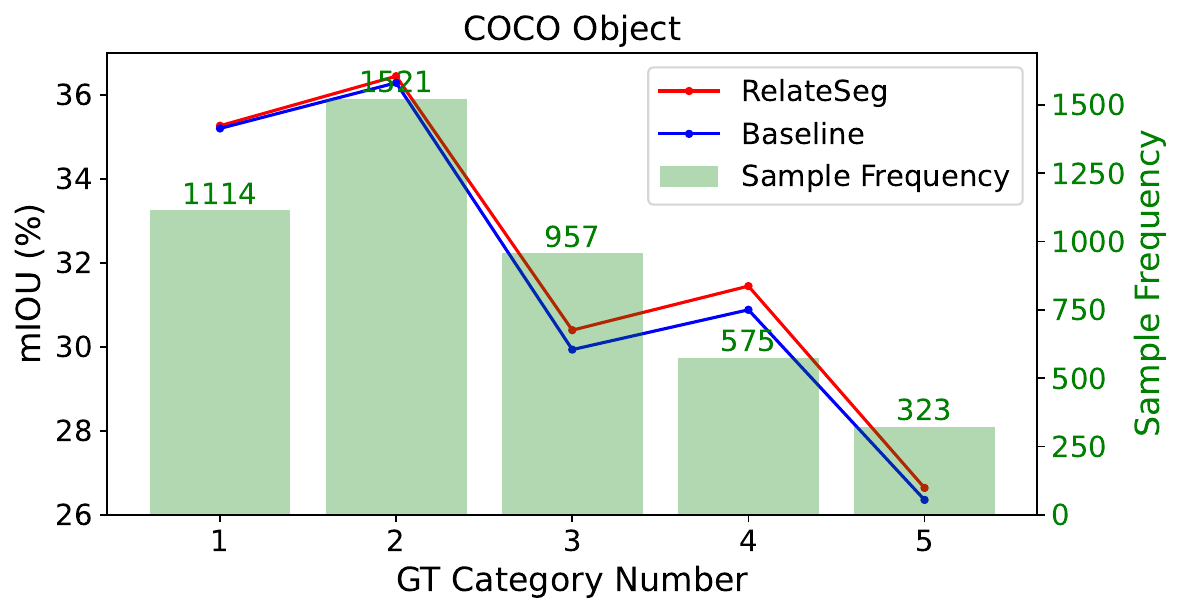}
    \end{subfigure}
  \caption{RelateSeg VS Baseline by category number per image on VOC12 and COCO.}\label{fig:delta_gt_cat}
\end{figure*}

\section{Experiments}
\subsection{Experimental Setup}
\textbf{Datasets and Metrics.} We evaluate \ours on four widely used open-vocabulary semantic segmentation (OVSS) benchmarks, namely, PASCAL VOC 2012(VOC12)~\cite{everingham2010pascal}, PASCAL Context~\cite{pascal0context}, COCO Object (COCO)~\cite{coco}  and ADE20K~\cite{ade}
containing 20, 59, 80 and 150 foreground categories and 1449, 5104,
5000, 500 and 2000 images, respectively in their validation sets.
Following prior works~\cite{diffseg, ovdiff, wang2025diffusion}, we use mean intersection over union (mIoU), while both mIoU and mean accuracy (mAcc) are used in our ablation study.


\noindent\textbf{Implementation Details.}
To obtain category names $\mathbf{C}$, we follow DiffSegmenter~\cite{wang2025diffusion} by using BLIP~\cite{li2022blip} and CLIP~\cite{clip}, and further complement with another VLM, AllSeeing~\cite{wang2023allseeing}.
We use llavaOV (7B variant)~\cite{li2024llava} to reason spatial relations.
We use InvSeg as the baseline~\cite{lin2025invseg}. Following the original InvSeg setup, each test image is augmented twice during the optimization process, resulting in a batch of size 2 at each optimization step. 
Different from InvSeg, we use random resized crops with minimum crop rate 0.8 instead of 0.6 to include multiple categories for applying spatial constraints.
We employ the Adam optimizer~\cite{Adam} with a learning rate of 0.01, optimizing 15 steps for each image.
The weight for spatial loss weight is 0.1.
\jy{We used {a} single H100 for all the experiments. }

\begin{figure}[t!]
\centering
 \includegraphics[width=.48\textwidth]{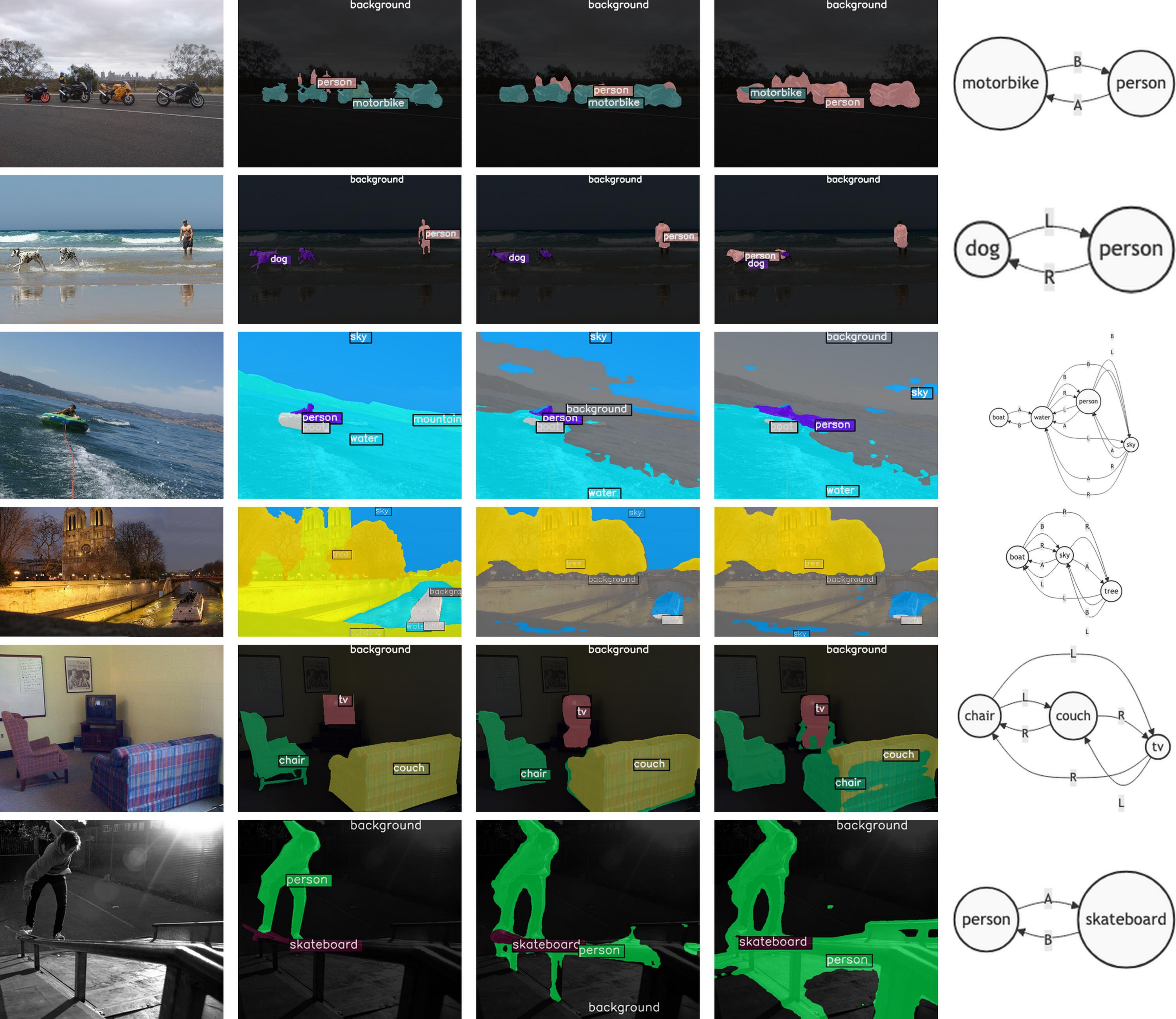  }   
  \caption{Examples of Segmentation on COCO (top), VOC12 (middle) and Context (bottom).
Each row (a group of five images) is organized as follows from left to right: input image, ground truth (GT), RelateSeg, InvSeg (our implementation), and the spatial graph. In the spatial graph, we use the following notations: 
  ``A'' = above,
  ``B'' = below,
  ``R'' = to the right of,
  ``L'' = to the left of.
  }
  \label{fig:sota1}

\end{figure}

\subsection{Comparative Evaluation on OVSS}
Table~\ref{tab:sota} shows a comparison between \ours and contemporary works on OVSS tasks.
As \ours does not rely on mask annotations or any additional training data, we primarily compare it with other mask-free methods, which fall into two categories:
(1) training-free methods, including:
CLIP~\cite{clip},
SCLIP~\cite{wang2024sclip},
CLIPSurgery~\cite{li2025closer},
MaskCLIP~\cite{maskclip}, DiffSegmenter~\cite{wang2025diffusion}, and our implementation of diffusion-based~\cite{Stable_diffusion} segmentation(\ours without test-time optimization).
(2) Test-time optimization methods, including:
InvSeg and Diffusion+TPT, which refers to our implementation of Test-time Prompt Tuning (TPT)~\cite{shu2022test} adapted for diffusion models.
We also list methods trained on extra image-text pairs for reference:
 TCL~\cite{tcl}, ViewCo~\cite{ren2023viewco},  SegCLIP~\cite{luo2023segclip}, and OVSegmentor~\cite{ovsegmenter}.
\ours is a test-time optimization method, and we primarily compare its performance with InvSeg and Diffusion+TPT.
On average across the four datasets, \ours achieves state-of-the-art performance, outperforming Diffusion+TPT.
Compared to InvSeg, \ours yields a 2.3\% mIoU improvement while maintaining competitive results on  VOC12 and COCO.

To further investigate where \ours excels, we reimplement InvSeg under the same settings as \ours to serve as a baseline.
We analyze the effect of spatial constraints on images with varying numbers of categories on VOC12 and COCO, as shown in Fig.~\ref{fig:delta_gt_cat}.
By grouping images based on their ground-truth category count, we observe that \ours outperforms InvSeg more and more obviously with increasing category count—particularly for images containing up to four categories.
At four categories, \ours achieves a 3.0\% mIoU gain on VOC12 and a 0.6\% gain on COCO over InvSeg.
These results demonstrate that spatial constraints become increasingly beneficial as scene complexity rises.
However, the more erratic performance beyond four categories highlights the challenges of over-complex scenes, such as overlapping objects and multiple instances per category.
These findings confirm both the effectiveness of spatial reasoning in OVSS and its current limitations in highly complex scenarios—offering avenues for future research.
The qualitative comparison is shown in Fig.~\ref{fig:sota1}.


\begin{table}[ht]
        \centering
        \setlength{\tabcolsep}{10pt}
        \resizebox{.45\textwidth}{!}{%
          \begin{tabular}{lccccc}
            \toprule
            \multicolumn{2}{l}{{\centering{Method's Variants}}} 
            & \multicolumn{2}{c}{\centering{PASCAL VOC}}
            & \multicolumn{2}{c}{\centering{COCO Object}} \\
            \midrule
            {$\alpha$}  & re-weight
            & mIOU & mAcc 
            & mIOU & mAcc \\
            \midrule
            \midrule
            0     &  -     & 53.7 & 73.3 & 30.3 & 49.1 \\
                    &&\textcolor{gray}{(62.8)} & \textcolor{gray}{(79.6)} & \textcolor{gray}{(35.0)} & \textcolor{gray}{(54.6)} \\
            0.1   &  \ding{55}    & 54.5 & 73.8 & 30.6 & 50.2 \\
                    &&\textcolor{gray}{(63.0)} & \textcolor{gray}{(79.6)} & \textcolor{gray}{(35.2)} & \textcolor{gray}{(55.2)} \\
            0.1   &  \checkmark   &\textbf{54.8} & \textbf{74.2} & \textbf{30.7}& \textbf{50.3} \\
                    &&\textcolor{gray}{(\textbf{63.1})} & \textcolor{gray}{(\textbf{79.7})} & \textcolor{gray}{(\textbf{35.3})} & \textcolor{gray}{(\textbf{55.3})} \\
            0.05   &  \checkmark  & 54.2 & 73.8 & 30.5 & 49.9 \\
                    &&\textcolor{gray}{(62.9)} & \textcolor{gray}{(79.6)} & \textcolor{gray}{(35.2)} & \textcolor{gray}{(55.0)} \\
            0.5   &  \checkmark  & 52.1 & 70.1 & 30.0 & 49.4 \\
                    &&\textcolor{gray}{(62.3)} & \textcolor{gray}{(77.9)} & \textcolor{gray}{(34.7)} & \textcolor{gray}{(54.8)} \\
            \bottomrule
          \end{tabular}%
        }
        \caption{Effect of spatial loss weight ($\alpha$) and constraint-specific re-weighting, evaluated on spatially constrained samples. \textcolor{gray}{Gray numbers} indicate performance on the full dataset.
}\label{tab:sk_loss}    
\end{table}
\begin{table}[ht]
        \centering
        
        \setlength{\tabcolsep}{5pt}
        \resizebox{.45\textwidth}{!}{%
          \begin{tabular}{lcccccc}
            \toprule
            \multicolumn{3}{l}{{\centering{Method's Variants}}} 
            & \multicolumn{2}{c}{\centering{PASCAL VOC}}
            & \multicolumn{2}{c}{\centering{COCO Object}} \\
            \midrule
            Polar & Contra & VLM
            & mIOU & mAcc 
            & mIOU & mAcc \\
            \midrule
            \midrule
                \ding{55}  & \ding{55}  &  LLaVA-OV7B  & \underline{63.1} & 79.4 & \underline{35.2} & 54.8\\
                \checkmark & \ding{55}  &  LLaVA-OV7B  & 63.1 & \underline{79.6} & 35.1 & \underline{54.9}\\
                \checkmark &  \checkmark &  LLaVA-OV7B & \textbf{63.1} & \textbf{79.7}& \textbf{35.3} & \textbf{55.3} \\ 
                \checkmark &  \checkmark &  LLaVA1.5   & 62.7 & 79.4 & 35.0 & 54.7\\
                \checkmark &  \checkmark &   LLaVA-OV  & 62.9 & 79.4 & 34.9 & 54.4\\
            \bottomrule
          \end{tabular}%
        }
\caption{Effect of applying different spatial rule filtering techniques and VLM models.
        Notation:
        ``Polar'': use polar questions to validate the triplets.
        ``Contra'': filter contradictory triplet pairs.}\label{tab:filter}

\end{table}
\begin{table}[ht]
    \centering
    \setlength{\tabcolsep}{5pt}
    \resizebox{0.47\textwidth}{!}{
    \begin{tabular}{lccc}
    \toprule
    Phase & VLM & Time / image & Memory \\
    \midrule
    \midrule
    {Compute} category names & CLIP, BLIP, AllSeeing & 15.0s & 32.9G \\
    \rowcolor{gray!10} {Compute} relations & llavaOV(7B) & 14.1s & \, 2.2G \\
    Test-time {optimization} & Stable Diffusion v2.1 & \; 7.2s & 32.0G \\
    \bottomrule
    \end{tabular}}
    \caption{{Computational costs} and Resource {breakdowns}}\label{tab:resource}
    
\end{table}
\setlength{\fboxsep}{1pt}
\begin{figure*}[t!]
\centering
    \begin{subfigure}
        \centering
        \includegraphics[width=0.42\textwidth]{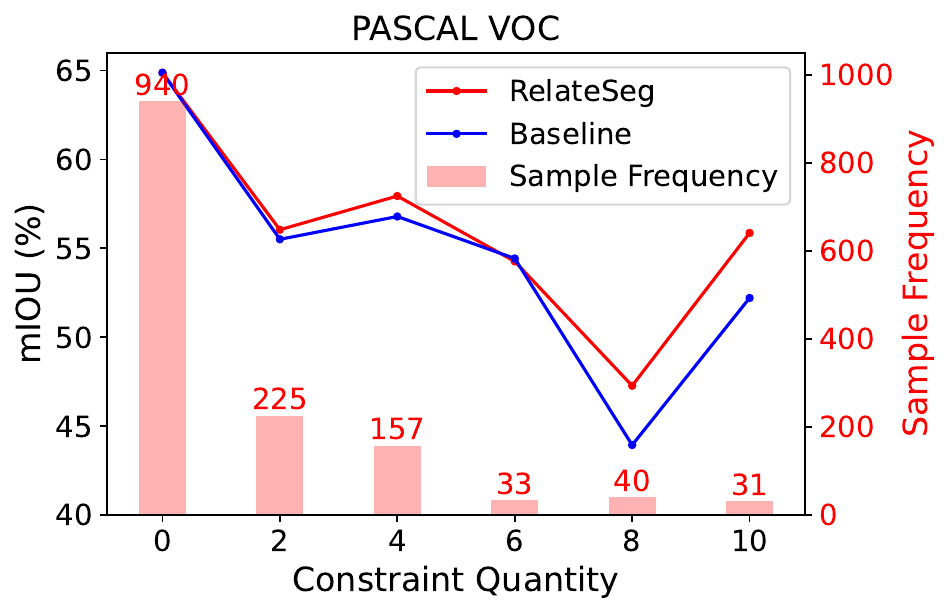}
    \end{subfigure}
    \hspace{.3cm}
    \begin{subfigure}
        \centering
        \includegraphics[width=0.533\textwidth]{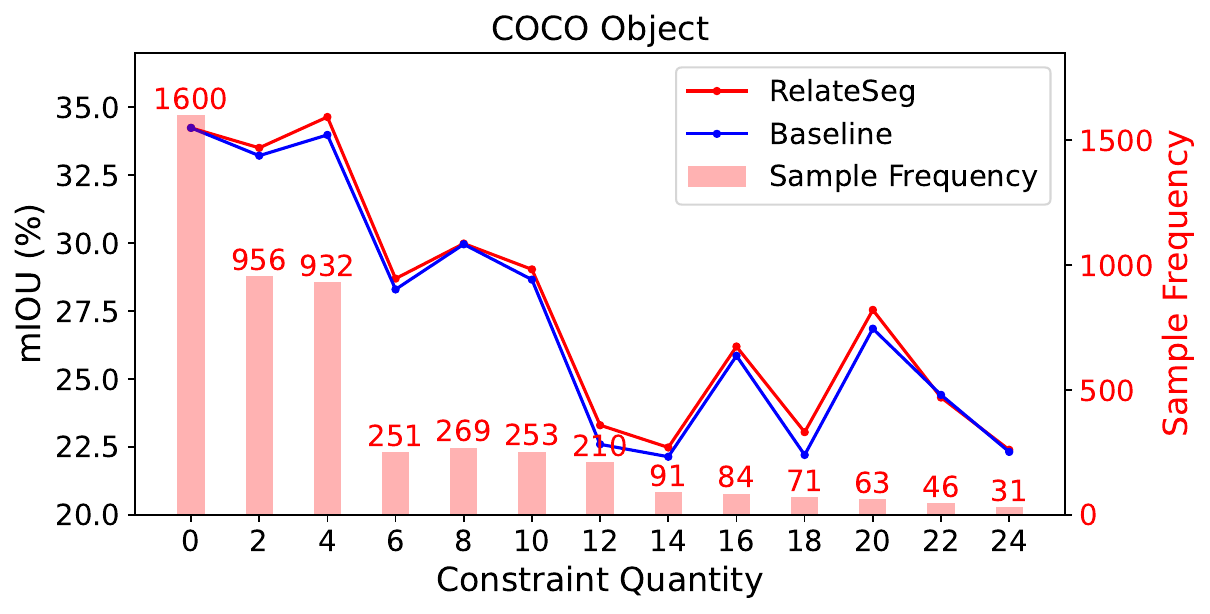}
    \end{subfigure}
    
    \begin{subfigure}
        \centering
        \includegraphics[width=0.42\textwidth]{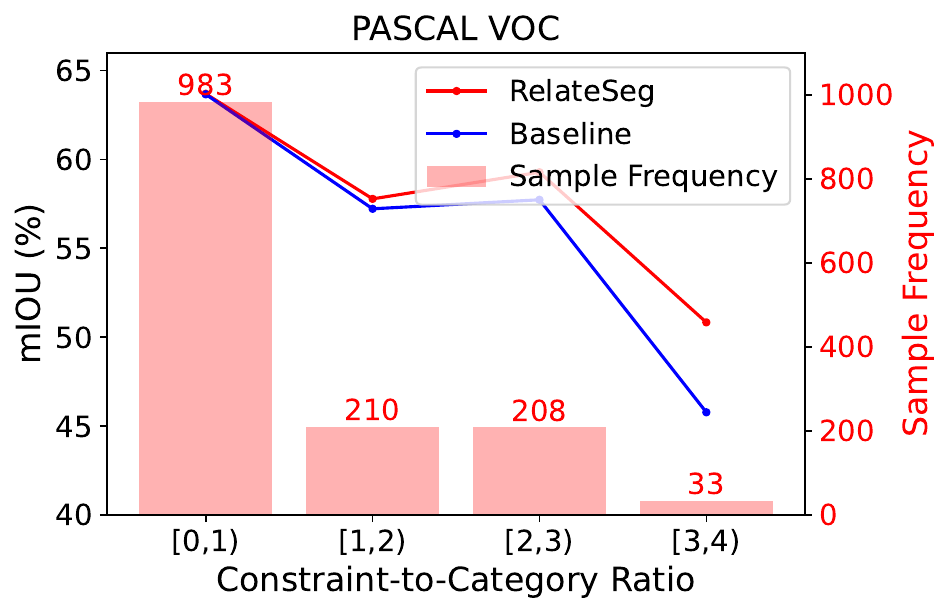}
    \end{subfigure}
    \hspace{.3cm}
    \begin{subfigure}
        \centering
        \includegraphics[width=0.533\textwidth]{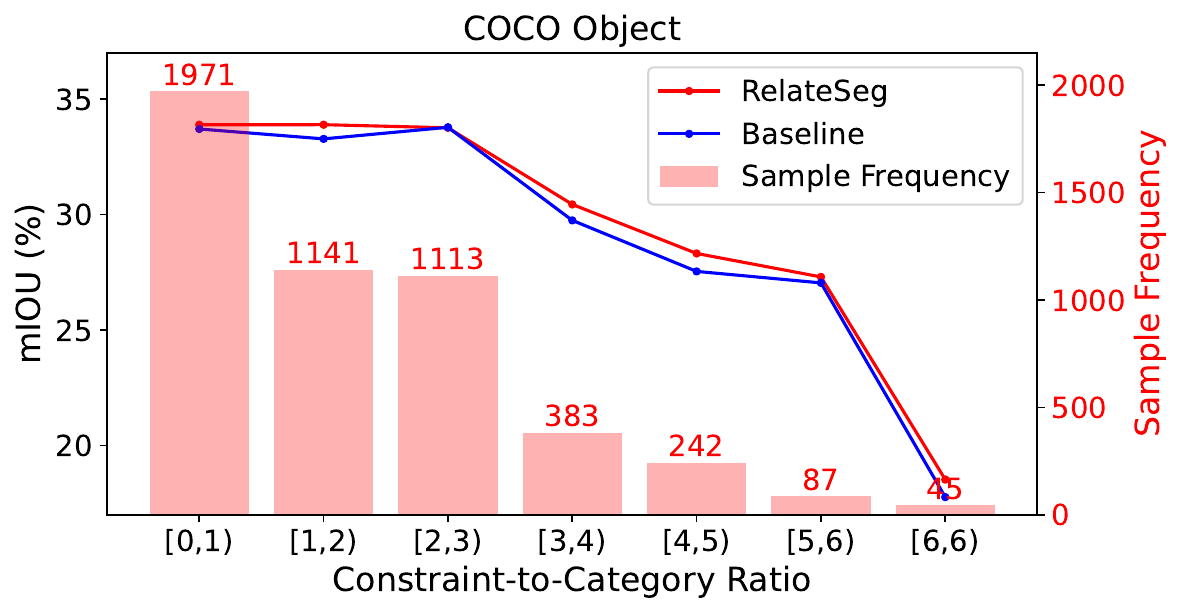}
    \end{subfigure}

  \caption{RelateSeg VS Baseline in mIOU(\%) across constraint Quantity (upper section) and constraint-to-category ratio (lower section) per image on VOC12 and COCO.}\label{fig:delta_rule}

\end{figure*}
\setlength{\fboxsep}{1pt}

\subsection{Ablation Studies}
We conduct a comprehensive analysis of various parameters and strategies during acquiring and applying spatial constraints on \ours.
Our experiments are carried out on PASCAL VOC 2012 and COCO Object datasets with metrics: mIOU and mAcc.
We study the effects of spatial constraint loss weight $\alpha$ and constraint-specific re-weighting in spatial loss computation in Tab~\ref{tab:sk_loss}.
We analyze only images with spatial constraints, as many images lack these constraints due to VLM limitations and the existence single-category images.
The first row shows the baseline with $\alpha=0$.
When $\alpha$ increases from 0 to 0.1, which means applying spatial constraints, VOC12 improves by 0.8\% mIOU and COCO improves by 0.3\% mIOU.
When performing constraint-specific re-weighting, we observe further improvements.
We find that $\alpha=0.1$ yields the best performance, while larger or smaller values of $\alpha$ degrade the results.
Furthermore, we study the effect of applying different spatial rule filtering techniques and different VLM models in Tab.~\ref{tab:filter}.
We observe that applying polar question validation and removing contradictory constraints leads to improvements in both mIOU and mAcc. 
Experimenting with different VLM models, including LLaVA1.5~\cite{liu2023visual, liu2024improved}, LLaVA-OV~\cite{li2024llava} and its 7B parameter variant yields consistently stable results.

\subsection{Further Analysis}

\jy{
\textbf{Computational Resources.}
Tab.~\ref{tab:resource} summarizes the VLMs, processing time, and memory usage for each phrase.
The first two phrases, \textit{Compute category names} and \textit{Compute relations}, prepare symbolic spatial constraints. The third phrase applies spatial reasoning in optimization, introducing only a new loss without additional time or memory overhead compared to InvSeg in practice.
\textit{Compute category names} is the most resource-intensive step. Test-time optimization is faster but remains memory-heavy. The main additional cost of \ours lies in \textit{Compute relations} (gray), which is relatively lightweight (14.1s, 2.2 GB), especially in memory usage.

}

\begin{figure}[t!]
\centering
\includegraphics[width=0.48\textwidth]{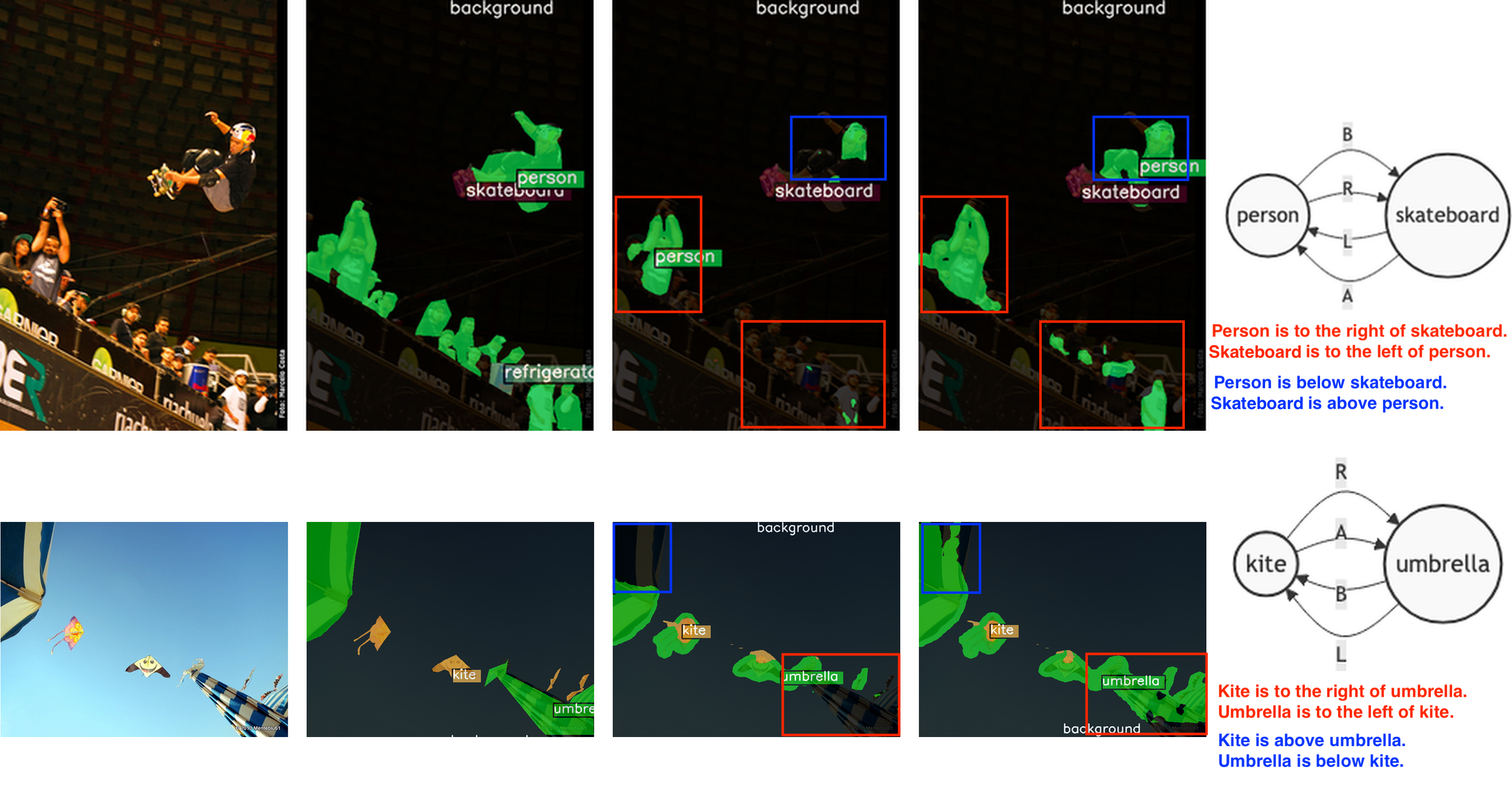}

\caption{Failure cases. 
Some parts of the segmentation masks—highlighted within bounding boxes—are negatively affected by specific spatial rules. The rules are indicated using the same color as the corresponding bounding boxes.}\label{fig:fail_case}

\end{figure}

\noindent\textbf{The effect of constraint number.}
We analyze the correlation between constraint quantity and segmentation performance to assess if more rules consistently improve performance (Fig.~\ref{fig:delta_rule}). 
The upper section of the figure shows that more spatial constraints initially improve segmentation quality (2 to 4 constraints), but the benefit becomes inconsistent beyond 6 constraints. 
To eliminate the effect of different category number in different images, the lower section examines the Constraint-to-Category Ratio (constraints per predicted category), 
when ratio in [3,4), the spatial constraints brings 5.1 \%, 0.7\% mIoU improvement on VOC12 and COCO, respectively. The performance benefits are consistent on VOC12 as the ratio increases, whereas COCO exhibits greater variability, though still demonstrating overall improvement.

\noindent\textbf{Failure cases.}
We show failure cases  in Fig.~\ref{fig:fail_case}. RelateSeg performs worse than InvSeg (the baseline) due to the ambiguity that arises when referring to multi-object categories. In such cases, certain spatial rules may not be consistently applicable to all instances of a category.

\section{Conclusion}
\jy{In this work, we presented RelateSeg, enhancing open-vocabulary semantic segmentation by introducing neuro-symbolic spatial reasoning. 
Our approach addresses the limitation of current vision-language models in understanding visual structure by learning spatial-relationally neuro-symbolic representation through proposed pseudo category.
Experiments across four benchmarks demonstrate that explicit spatial reasoning notably improves segmentation.}

\section{Limitations}
\ours faces challenges in capturing complex spatial dependencies, especially with many semantic categories or spatially dispersed instances. It also inherits VLM limitations in extracting reliable spatial cues from crowded scenes and struggles with representing certain relations in 2D space. Additionally, the method has relatively slow inference. Future work includes improving efficiency via knowledge distillation and caching of spatial relations.

\bibliography{main}

\begin{thebibliography}{53}
\providecommand{\natexlab}[1]{#1}

\bibitem[{Arbelaez et~al.(2010)Arbelaez, Maire, Fowlkes, and Malik}]{arbelaez2010contour}
Arbelaez, P.; Maire, M.; Fowlkes, C.; and Malik, J. 2010.
\newblock Contour detection and hierarchical image segmentation.
\newblock \emph{IEEE transactions on pattern analysis and machine intelligence}, 33(5): 898--916.

\bibitem[{Bertinetto et~al.(2020)Bertinetto, Mueller, Tertikas, Samangooei, and Lord}]{bertinetto2020making}
Bertinetto, L.; Mueller, R.; Tertikas, K.; Samangooei, S.; and Lord, N.~A. 2020.
\newblock Making better mistakes: Leveraging class hierarchies with deep networks.
\newblock In \emph{Proceedings of the IEEE/CVF conference on computer vision and pattern recognition}, 12506--12515.

\bibitem[{Cha, Mun, and Roh(2023)}]{tcl}
Cha, J.; Mun, J.; and Roh, B. 2023.
\newblock Learning to generate text-grounded mask for open-world semantic segmentation from only image-text pairs.
\newblock In \emph{Proceedings of the IEEE/CVF Conference on Computer Vision and Pattern Recognition}, 11165--11174.

\bibitem[{Chen et~al.(2020)Chen, Li, Huang, Zhang, and Ma}]{chen2020improving}
Chen, S.; Li, Z.; Huang, F.; Zhang, C.; and Ma, H. 2020.
\newblock Improving object detection with relation mining network.
\newblock In \emph{2020 IEEE International Conference on Data Mining (ICDM)}, 52--61. IEEE.

\bibitem[{Evans and Stanovich(2013)}]{evans2013dual}
Evans, J. S.~B.; and Stanovich, K.~E. 2013.
\newblock Dual-process theories of higher cognition: Advancing the debate.
\newblock \emph{Perspectives on psychological science}, 8(3): 223--241.

\bibitem[{Everingham et~al.(2010)Everingham, Van~Gool, Williams, Winn, and Zisserman}]{everingham2010pascal}
Everingham, M.; Van~Gool, L.; Williams, C.~K.; Winn, J.; and Zisserman, A. 2010.
\newblock The pascal visual object classes (voc) challenge.
\newblock \emph{International journal of computer vision}, 88(2): 303--338.

\bibitem[{Feldman(2003)}]{feldman2003visual}
Feldman, J. 2003.
\newblock What is a visual object?
\newblock \emph{Trends in Cognitive Sciences}, 7(6): 252--256.

\bibitem[{Giunchiglia and Lukasiewicz(2020)}]{giunchiglia2020coherent}
Giunchiglia, E.; and Lukasiewicz, T. 2020.
\newblock Coherent hierarchical multi-label classification networks.
\newblock \emph{Advances in neural information processing systems}, 33: 9662--9673.

\bibitem[{Giunchiglia and Lukasiewicz(2021)}]{giunchiglia2021multi}
Giunchiglia, E.; and Lukasiewicz, T. 2021.
\newblock Multi-label classification neural networks with hard logical constraints.
\newblock \emph{Journal of Artificial Intelligence Research}, 72: 759--818.

\bibitem[{H{\'a}jek(2013)}]{hajek2013metamathematics}
H{\'a}jek, P. 2013.
\newblock \emph{Metamathematics of fuzzy logic}, volume~4.
\newblock Springer Science \& Business Media.

\bibitem[{Kahneman(2011)}]{kahneman2011thinking}
Kahneman, D. 2011.
\newblock \emph{Thinking, fast and slow}.
\newblock macmillan.

\bibitem[{Karazija et~al.(2023)Karazija, Laina, Vedaldi, and Rupprecht}]{ovdiff}
Karazija, L.; Laina, I.; Vedaldi, A.; and Rupprecht, C. 2023.
\newblock Diffusion Models for Zero-Shot Open-Vocabulary Segmentation.
\newblock \emph{arXiv preprint arXiv:2306.09316}.

\bibitem[{Kim, Jung, and Lee(2021)}]{kim2021spatial}
Kim, G.; Jung, H.-G.; and Lee, S.-W. 2021.
\newblock Spatial reasoning for few-shot object detection.
\newblock \emph{Pattern Recognition}, 120: 108118.

\bibitem[{Kingma and Ba(2015)}]{Adam}
Kingma, D.~P.; and Ba, J. 2015.
\newblock Adam: {A} Method for Stochastic Optimization.
\newblock In Bengio, Y.; and LeCun, Y., eds., \emph{3rd International Conference on Learning Representations, {ICLR} 2015, San Diego, CA, USA, May 7-9, 2015, Conference Track Proceedings}.

\bibitem[{Lake et~al.(2017)Lake, Ullman, Tenenbaum, and Gershman}]{lake2017building}
Lake, B.~M.; Ullman, T.~D.; Tenenbaum, J.~B.; and Gershman, S.~J. 2017.
\newblock Building machines that learn and think like people.
\newblock \emph{Behavioral and brain sciences}, 40: e253.

\bibitem[{Li et~al.(2024)Li, Zhang, Guo, Zhang, Li, Zhang, Zhang, Zhang, Li, Liu et~al.}]{li2024llava}
Li, B.; Zhang, Y.; Guo, D.; Zhang, R.; Li, F.; Zhang, H.; Zhang, K.; Zhang, P.; Li, Y.; Liu, Z.; et~al. 2024.
\newblock Llava-onevision: Easy visual task transfer.
\newblock \emph{arXiv preprint arXiv:2408.03326}.

\bibitem[{Li et~al.(2022{\natexlab{a}})Li, Li, Xiong, and Hoi}]{li2022blip}
Li, J.; Li, D.; Xiong, C.; and Hoi, S. 2022{\natexlab{a}}.
\newblock Blip: Bootstrapping language-image pre-training for unified vision-language understanding and generation.
\newblock In \emph{International conference on machine learning}, 12888--12900. PMLR.

\bibitem[{Li, Wang, and Yang(2023)}]{li2023logicseg}
Li, L.; Wang, W.; and Yang, Y. 2023.
\newblock Logicseg: Parsing visual semantics with neural logic learning and reasoning.
\newblock In \emph{Proceedings of the IEEE/CVF International Conference on Computer Vision}, 4122--4133.

\bibitem[{Li et~al.(2022{\natexlab{b}})Li, Zhou, Wang, Li, and Yang}]{li2022deep}
Li, L.; Zhou, T.; Wang, W.; Li, J.; and Yang, Y. 2022{\natexlab{b}}.
\newblock Deep hierarchical semantic segmentation.
\newblock In \emph{Proceedings of the IEEE/CVF Conference on Computer Vision and Pattern Recognition}, 1246--1257.

\bibitem[{Li et~al.(2025)Li, Wang, Duan, Zhang, and Li}]{li2025closer}
Li, Y.; Wang, H.; Duan, Y.; Zhang, J.; and Li, X. 2025.
\newblock A closer look at the explainability of contrastive language-image pre-training.
\newblock \emph{Pattern Recognition}, 111409.

\bibitem[{Li et~al.(2023)Li, Zhou, Zhang, Zhang, Wang, and Xie}]{li2023open}
Li, Z.; Zhou, Q.; Zhang, X.; Zhang, Y.; Wang, Y.; and Xie, W. 2023.
\newblock Open-vocabulary object segmentation with diffusion models.
\newblock In \emph{Proceedings of the IEEE/CVF International Conference on Computer Vision}, 7667--7676.

\bibitem[{Lin et~al.(2025)Lin, Huang, Hu, and Gong}]{lin2025invseg}
Lin, J.; Huang, J.; Hu, J.; and Gong, S. 2025.
\newblock InvSeg: Test-Time Prompt Inversion for Semantic Segmentation.
\newblock In \emph{Proceedings of the AAAI Conference on Artificial Intelligence}, 5245--5253.

\bibitem[{Lin et~al.(2014)Lin, Maire, Belongie, Hays, Perona, Ramanan, Doll{\'a}r, and Zitnick}]{coco}
Lin, T.-Y.; Maire, M.; Belongie, S.; Hays, J.; Perona, P.; Ramanan, D.; Doll{\'a}r, P.; and Zitnick, C.~L. 2014.
\newblock Microsoft coco: Common objects in context.
\newblock In \emph{Computer Vision--ECCV 2014: 13th European Conference, Zurich, Switzerland, September 6-12, 2014, Proceedings, Part V 13}, 740--755. Springer.

\bibitem[{Liu et~al.(2024)Liu, Li, Li, and Lee}]{liu2024improved}
Liu, H.; Li, C.; Li, Y.; and Lee, Y.~J. 2024.
\newblock Improved baselines with visual instruction tuning.
\newblock In \emph{Proceedings of the IEEE/CVF Conference on Computer Vision and Pattern Recognition}, 26296--26306.

\bibitem[{Liu et~al.(2023)Liu, Li, Wu, and Lee}]{liu2023visual}
Liu, H.; Li, C.; Wu, Q.; and Lee, Y.~J. 2023.
\newblock Visual instruction tuning.
\newblock \emph{arXiv preprint arXiv:2304.08485}.

\bibitem[{Luo et~al.(2023)Luo, Bao, Wu, He, and Li}]{luo2023segclip}
Luo, H.; Bao, J.; Wu, Y.; He, X.; and Li, T. 2023.
\newblock Segclip: Patch aggregation with learnable centers for open-vocabulary semantic segmentation.
\newblock In \emph{International Conference on Machine Learning}, 23033--23044. PMLR.

\bibitem[{Malik et~al.(2001)Malik, Belongie, Leung, and Shi}]{malik2001contour}
Malik, J.; Belongie, S.; Leung, T.; and Shi, J. 2001.
\newblock Contour and texture analysis for image segmentation.
\newblock \emph{International journal of computer vision}, 43: 7--27.

\bibitem[{Marcus(2018)}]{marcus2018deep}
Marcus, G. 2018.
\newblock Deep Learning: A Critical Appraisal.
\newblock \emph{arXiv preprint arXiv:1801.00631}.

\bibitem[{Marr and Nishihara(1978)}]{marr1978representation}
Marr, D.; and Nishihara, H.~K. 1978.
\newblock Representation and recognition of the spatial organization of three-dimensional shapes.
\newblock \emph{Proceedings of the Royal Society of London. Series B. Biological Sciences}, 200(1140): 269--294.

\bibitem[{McCulloch and Pitts(1943)}]{mcculloch1943logical}
McCulloch, W.~S.; and Pitts, W. 1943.
\newblock A logical calculus of the ideas immanent in nervous activity.
\newblock \emph{The bulletin of mathematical biophysics}, 5: 115--133.

\bibitem[{Mottaghi et~al.(2014)Mottaghi, Chen, Liu, Cho, Lee, Fidler, Urtasun, and Yuille}]{pascal0context}
Mottaghi, R.; Chen, X.; Liu, X.; Cho, N.-G.; Lee, S.-W.; Fidler, S.; Urtasun, R.; and Yuille, A. 2014.
\newblock The role of context for object detection and semantic segmentation in the wild.
\newblock In \emph{Proceedings of the IEEE conference on computer vision and pattern recognition}, 891--898.

\bibitem[{Prokopowicz et~al.(2017)Prokopowicz, Czerniak, Miko{\l}ajewski, Apiecionek, and {\'S}lezak}]{prokopowicz2017theory}
Prokopowicz, P.; Czerniak, J.; Miko{\l}ajewski, D.; Apiecionek, {\L}.; and {\'S}lezak, D. 2017.
\newblock \emph{Theory and applications of ordered fuzzy numbers: a tribute to Professor Witold Kosi{\'n}ski}.
\newblock Springer Nature.

\bibitem[{Radford et~al.(2021)Radford, Kim, Hallacy, Ramesh, Goh, Agarwal, Sastry, Askell, Mishkin, Clark et~al.}]{clip}
Radford, A.; Kim, J.~W.; Hallacy, C.; Ramesh, A.; Goh, G.; Agarwal, S.; Sastry, G.; Askell, A.; Mishkin, P.; Clark, J.; et~al. 2021.
\newblock Learning transferable visual models from natural language supervision.
\newblock In \emph{International conference on machine learning}, 8748--8763. PMLR.

\bibitem[{Ren et~al.(2023)Ren, Li, Xu, Zhu, Wang, Liu, Chang, and Liang}]{ren2023viewco}
Ren, P.; Li, C.; Xu, H.; Zhu, Y.; Wang, G.; Liu, J.; Chang, X.; and Liang, X. 2023.
\newblock ViewCo: Discovering Text-Supervised Segmentation Masks via Multi-View Semantic Consistency.
\newblock In \emph{The Eleventh International Conference on Learning Representations}.

\bibitem[{Rombach et~al.(2022)Rombach, Blattmann, Lorenz, Esser, and Ommer}]{Stable_diffusion}
Rombach, R.; Blattmann, A.; Lorenz, D.; Esser, P.; and Ommer, B. 2022.
\newblock High-resolution image synthesis with latent diffusion models.
\newblock In \emph{Proceedings of the IEEE/CVF conference on computer vision and pattern recognition}, 10684--10695.

\bibitem[{Shu et~al.(2022)Shu, Nie, Huang, Yu, Goldstein, Anandkumar, and Xiao}]{shu2022test}
Shu, M.; Nie, W.; Huang, D.-A.; Yu, Z.; Goldstein, T.; Anandkumar, A.; and Xiao, C. 2022.
\newblock Test-time prompt tuning for zero-shot generalization in vision-language models.
\newblock \emph{Advances in Neural Information Processing Systems}, 35: 14274--14289.

\bibitem[{Sueyoshi and Matsubara(2024)}]{sueyoshi2024predicated}
Sueyoshi, K.; and Matsubara, T. 2024.
\newblock Predicated Diffusion: Predicate Logic-Based Attention Guidance for Text-to-Image Diffusion Models.
\newblock In \emph{Proceedings of the IEEE/CVF Conference on Computer Vision and Pattern Recognition}, 8651--8660.

\bibitem[{Tian et~al.(2023)Tian, Aggarwal, Colaco, Kira, and Gonzalez-Franco}]{diffseg}
Tian, J.; Aggarwal, L.; Colaco, A.; Kira, Z.; and Gonzalez-Franco, M. 2023.
\newblock Diffuse, attend, and segment: Unsupervised zero-shot segmentation using stable diffusion.
\newblock \emph{arXiv preprint arXiv:2308.12469}.

\bibitem[{van Krieken, Acar, and van Harmelen(2022)}]{van2022analyzing}
van Krieken, E.; Acar, E.; and van Harmelen, F. 2022.
\newblock Analyzing differentiable fuzzy logic operators.
\newblock \emph{Artificial Intelligence}, 302: 103602.

\bibitem[{Wang, Mei, and Yuille(2024)}]{wang2024sclip}
Wang, F.; Mei, J.; and Yuille, A. 2024.
\newblock Sclip: Rethinking self-attention for dense vision-language inference.
\newblock In \emph{European Conference on Computer Vision}, 315--332. Springer.

\bibitem[{Wang et~al.(2025)Wang, Li, Zhang, Xu, Zhou, Yu, Sheng, and Xu}]{wang2025diffusion}
Wang, J.; Li, X.; Zhang, J.; Xu, Q.; Zhou, Q.; Yu, Q.; Sheng, L.; and Xu, D. 2025.
\newblock Diffusion Model is Secretly a Training-Free Open Vocabulary Semantic Segmenter.
\newblock \emph{IEEE Transactions on Image Processing}.

\bibitem[{Wang et~al.(2023)Wang, Shi, Li, Wang, Huang, Xing, Chen, Li, Zhu, Cao et~al.}]{wang2023allseeing}
Wang, W.; Shi, M.; Li, Q.; Wang, W.; Huang, Z.; Xing, L.; Chen, Z.; Li, H.; Zhu, X.; Cao, Z.; et~al. 2023.
\newblock The All-Seeing Project: Towards Panoptic Visual Recognition and Understanding of the Open World.
\newblock \emph{arXiv preprint arXiv:2308.01907}.

\bibitem[{Wang, Yang, and Wu(2024)}]{wang2024towards}
Wang, W.; Yang, Y.; and Wu, F. 2024.
\newblock Towards Data-and Knowledge-Driven AI: A Survey on Neuro-Symbolic Computing.
\newblock \emph{IEEE Transactions on Pattern Analysis and Machine Intelligence}.

\bibitem[{Wang et~al.(2019)Wang, Zhang, Qi, Shen, Pang, and Shao}]{wang2019learning}
Wang, W.; Zhang, Z.; Qi, S.; Shen, J.; Pang, Y.; and Shao, L. 2019.
\newblock Learning compositional neural information fusion for human parsing.
\newblock In \emph{Proceedings of the IEEE/CVF international conference on computer vision}, 5703--5713.

\bibitem[{Wang et~al.(2024)Wang, Sun, Luo, Pan, and Zhang}]{wang2024image}
Wang, Y.; Sun, R.; Luo, N.; Pan, Y.; and Zhang, T. 2024.
\newblock Image-to-Image Matching via Foundation Models: A New Perspective for Open-Vocabulary Semantic Segmentation.
\newblock In \emph{Proceedings of the IEEE/CVF Conference on Computer Vision and Pattern Recognition}, 3952--3963.

\bibitem[{Wehrmann, Cerri, and Barros(2018)}]{wehrmann2018hierarchical}
Wehrmann, J.; Cerri, R.; and Barros, R. 2018.
\newblock Hierarchical multi-label classification networks.
\newblock In \emph{International conference on machine learning}, 5075--5084. PMLR.

\bibitem[{Wu et~al.(2024{\natexlab{a}})Wu, Chen, Ji, Wang, Ma, Huang, Luo, Fei, Sun, and Ji}]{wu2024rgsan}
Wu, C.; Chen, Q.; Ji, J.; Wang, H.; Ma, Y.; Huang, Y.; Luo, G.; Fei, H.; Sun, X.; and Ji, R. 2024{\natexlab{a}}.
\newblock {RG}-{SAN}: Rule-Guided Spatial Awareness Network for End-to-End 3D Referring Expression Segmentation.
\newblock In \emph{The Thirty-eighth Annual Conference on Neural Information Processing Systems}.

\bibitem[{Wu et~al.(2024{\natexlab{b}})Wu, Chang, Chuang, Chen, Liu, Chen, Hu, Chuang, and Lin}]{wu2024image}
Wu, J.-J.; Chang, A. C.-H.; Chuang, C.-Y.; Chen, C.-P.; Liu, Y.-L.; Chen, M.-H.; Hu, H.-N.; Chuang, Y.-Y.; and Lin, Y.-Y. 2024{\natexlab{b}}.
\newblock Image-Text Co-Decomposition for Text-Supervised Semantic Segmentation.
\newblock In \emph{Proceedings of the IEEE/CVF Conference on Computer Vision and Pattern Recognition}, 26794--26803.

\bibitem[{Wu et~al.(2021)Wu, Wang, Hou, Lin, and Luo}]{wu2021spatial}
Wu, X.; Wang, R.; Hou, J.; Lin, H.; and Luo, J. 2021.
\newblock Spatial--temporal relation reasoning for action prediction in videos.
\newblock \emph{International Journal of Computer Vision}, 129(5): 1484--1505.

\bibitem[{Wysocza{\'n}ska et~al.(2024)Wysocza{\'n}ska, Sim{\'e}oni, Ramamonjisoa, Bursuc, Trzci{\'n}ski, and P{\'e}rez}]{wysoczanska2024clip}
Wysocza{\'n}ska, M.; Sim{\'e}oni, O.; Ramamonjisoa, M.; Bursuc, A.; Trzci{\'n}ski, T.; and P{\'e}rez, P. 2024.
\newblock CLIP-DINOiser: Teaching CLIP a few DINO tricks for open-vocabulary semantic segmentation.
\newblock In \emph{European Conference on Computer Vision}, 320--337. Springer.

\bibitem[{Xu et~al.(2023)Xu, Hou, Zhang, Feng, Wang, Qiao, and Xie}]{ovsegmenter}
Xu, J.; Hou, J.; Zhang, Y.; Feng, R.; Wang, Y.; Qiao, Y.; and Xie, W. 2023.
\newblock Learning open-vocabulary semantic segmentation models from natural language supervision.
\newblock In \emph{Proceedings of the IEEE/CVF Conference on Computer Vision and Pattern Recognition}, 2935--2944.

\bibitem[{Zhou et~al.(2019)Zhou, Zhao, Puig, Xiao, Fidler, Barriuso, and Torralba}]{ade}
Zhou, B.; Zhao, H.; Puig, X.; Xiao, T.; Fidler, S.; Barriuso, A.; and Torralba, A. 2019.
\newblock Semantic understanding of scenes through the ade20k dataset.
\newblock \emph{International Journal of Computer Vision}, 127: 302--321.

\bibitem[{Zhou, Loy, and Dai(2022)}]{maskclip}
Zhou, C.; Loy, C.~C.; and Dai, B. 2022.
\newblock Extract free dense labels from clip.
\newblock In \emph{European Conference on Computer Vision}, 696--712. Springer.

\end{thebibliography}

\newpage






\title{
Technical Appendices and Supplementary Material for\\
``Neuro-Symbolic Spatial Reasoning in Segmentation''
}
%

\author{%
  David S.~Hippocampus\thanks{Use footnote for providing further information
    about author (webpage, alternative address)---\emph{not} for acknowledging
    funding agencies.} \\
  Department of Computer Science\\
  Cranberry-Lemon University\\
  Pittsburgh, PA 15213 \\
  \texttt{hippo@cs.cranberry-lemon.edu} \\
}




\section{Correctness of extracted categories through VLMs}
The accuracy of category extraction through Vision-Language Models (VLMs) is crucial for segmentation performance. These extracted categories serve as candidate classes for segmentation and form the foundation for inferring spatial relationships between category pairs.

Table~\ref{tab:category_acc} presents the evaluation results across different datasets. PASCAL VOC achieves the highest F1 score of 92.3\%, demonstrating excellent category extraction performance. Both Average Precision (AP) and Average Recall (AR) exhibit a general correlation with dataset complexity and category count. For instance, more complex datasets like ADE20k, which contains a larger number of fine-grained categories, show lower performance metrics, with an F1 score of 53.0\%. This pattern reflects the inherent challenge of category extraction in datasets with extensive category vocabularies and semantic diversity.

\begin{table}[h]
\caption{Correctness of categories through VLMs.}
\label{tab:category_acc}
\centering
\setlength{\tabcolsep}{5pt}
\resizebox{0.46\textwidth}{!}{
\begin{tabular}{lcccc}
\toprule
Metric & PASCAL VOC & PASCAL Context & COCO Object  & ADE20k \\
\midrule
\midrule
Average Precision   & 88.7 & 72.6 & 81.2 & 61.5 \\
Average Recall      & 96.2 & 69.3 & 82.4 & 46.6 \\
F1 Score            & 92.3 & 70.9 & 81.8 & 53.0 \\
\bottomrule
\end{tabular}}
\end{table}



\section{Simpler Relation Set}
We conducted the ablation studies on relation subsets on COCO dataset, where only {“above”, “below”} or {“left”, “right”} in Tab.~\ref{tab:sk_simple}.These results indicate that horizontal relations contribute most significantly to performance gains, while the complete set yields better overall accuracy. This suggests that while simpler relation sets can still be useful, the full directional set provides optimal performance. 

\begin{table}[h]
\caption{Effect of different types of relations.
}\label{tab:sk_simple}
\centering
\setlength{\tabcolsep}{25pt}
\resizebox{0.46\textwidth}{!}{
  \begin{tabular}{c|cc}
    \toprule
    \multicolumn{1}{c}{{\centering{Method's Variants}}} 
    & \multicolumn{2}{c}{\centering{COCO Object}} \\
    \midrule
    Relations & mIOU & mAcc \\
    \midrule
    \midrule
    None & 30.3 & 49.1 \\
     & \gray{(35.0)} & \gray{(54.6)} \\
    Above-Below &  30.2 & 49.8 \\
     & \gray{(35.0)} & \gray{(55.0)} \\
    Left-Right &  \textbf{30.8} & 50.1 \\
     & \gray{(35.3)} & \gray{(55.1)} \\
    All &  30.7 & \textbf{50.3} \\
     & \gray{(35.3)} & \gray{(55.3)} \\
    \bottomrule
   \end{tabular}}
\end{table}

\section{Implementation Details of Spatial Knowledge Acquisition}

\begin{figure}[H]
    \centering
    \includegraphics[width=0.5\linewidth]{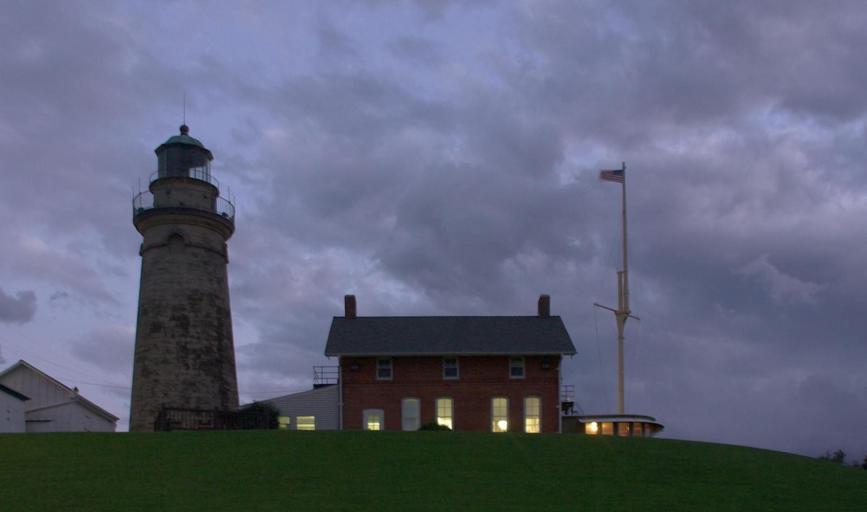}
    \caption{Example image.}
    \label{fig:example_img}
\end{figure}

This is the implementation details of ``Sec. 3.1 Spatial Knowledge Acquisition'' in the main text.
We extract spatial relation triplets in the format $\langle \text{subject}, \text{relation}, \text{object} \rangle$ from images using Vision-Language Models (VLMs) based on the list of categories  
$\mathbf{C} = \{c_i \mid i = 1, 2, \ldots, C\}$ presented in the image. 
The process includes augmentation and validation.
We illustrate this process with an example image in Fig.~\ref{fig:example_img}, with $\mathbf{C}=\{\text{``building''}, \text{``sky''}, \text{``tree''}, \text{``grass''}, \text{``house''}, \text{``hill''}, \text{``flag''}\}$.

\subsection{Caption Generation}
Generate descriptive caption of the image to provide context for spatial relation questions.
\begin{lstlisting}
Q: <image>. Describe this image in one sentence.
A: A lighthouse and a house with lights on are on a hill under a cloudy sky.
\end{lstlisting}

\subsection{Initial Triplet Generation}
For each pair of categories, based on the caption, we ask spatial relation questions using a specific prompt format using
  a predefined set of fundamental positional relations:  
$\mathbf{R} = \{\text{``above''}, \text{``below''}, \text{``left''}, \text{``right''}\}$.

\textbf{Example prompts from the output:}
\begin{lstlisting}
Q0: <image>. A lighthouse and a house with lights on are on a hill under a cloudy sky. From the perspective of someone looking at the image, is the building positioned above or below the sky?
A0: Below

Q0: <image>. A lighthouse and a house with lights on are on a hill under a cloudy sky. From the perspective of someone looking at the image, is the building positioned to the left or right of the sky?
A0: above
\end{lstlisting}

In this way, we extract the initial triplets (81 total):
\begin{lstlisting}
[["sky", "bottom", "building"], 
 ["sky", "right", "building"], 
 ["tree", "bottom", "building"], 
 ["grass", "top", "building"],
 ["house", "bottom", "building"],
 ["flag", "left", "building"],
 ...]
\end{lstlisting}

\subsection{Bidirectional Augmentation}
We generate reverse relations to complement the triplet lists.

\textbf{Example:}
\begin{itemize}
    \item Original: $\langle \text{"sky"}, \text{"bottom"}, \text{"building"} \rangle$
    \item Bidirectional: $\langle \text{"building"}, \text{"top"}, \text{"sky"} \rangle$
\end{itemize}
After Bidirectional Augmentation. We obtian 113 triplets in total.

\subsection{Polar Question Validation}
To ensure bidirectional consistency, we filter out spatial relations that work in both directions using following validation prompt templates:

{Primary Validation Question Format:}
\begin{lstlisting}
{caption}. From the perspective of someone looking at the image, 
is there a {object} {preposition} a {subject}?
\end{lstlisting}

{Reflection Validation Question Format:}
\begin{lstlisting}
{caption}. From the perspective of someone looking at the image, 
is there a {subject} {opposite_preposition} a {object}?
\end{lstlisting}

\textbf{Example:}

For triplet $\langle \text{"building"}, \text{"bottom"}, \text{"sky"} \rangle$

{Primary Question:}
\begin{lstlisting}
Q: <image>
A lighthouse and a house with lights on are on a hill under a cloudy sky. 
From the perspective of someone looking at the image, is there a sky below a building?
A: Yes
\end{lstlisting}

{Reflection Question:}
\begin{lstlisting}
Q: <image>
A lighthouse and a house with lights on are on a hill under a cloudy sky. 
From the perspective of someone looking at the image, is there a building above a sky?
A: Yes
\end{lstlisting}

\textbf{Validation Logic:}
\begin{align}
\text{Result} &= \text{Primary} \land \text{Reflection} \\
&= \text{True} \land \text{True} \\
&= \text{True} \quad \rightarrow \text{VALIDATED}
\end{align}

Both primary and reflection questions must return \texttt{True} for validation.
After validation of 113 bidirectional triplets, 52 triplets (46\% validation rate) are selected.

\subsection{Contradiction Filtering:}
In this process, we consider all triplets jointly to ensure their mutual consistency, rather than evaluating them individually as done previously. We apply spatial logic rules to identify the contradictory relations.

\textbf{Spatial Logic Rules Applied:}
\begin{enumerate}
    \item \textbf{Mutual Exclusivity}: Object cannot be both "left" and "right" of same target
    \item \textbf{Mutual Exclusivity}: Object cannot be both "top" and "bottom" of same target  
    \item \textbf{Bidirectional Consistency}: Same relation in both directions is impossible
\end{enumerate}

\subsubsection{Detect Contradictions}
We analyze category pairs to identify two types of spatial contradictions. 

\textbf{Contradiction Type 1: Cyclic Contradiction}

This type occurs when both objects claim the same spatial relationship to each other, creating a logical impossibility.

\begin{align}
\text{Triplet 1:} &\quad \langle \text{"building"}, \text{"right"}, \text{"sky"} \rangle \\
\text{Triplet 2:} &\quad \langle \text{"sky"}, \text{"left"}, \text{"building"} \rangle
\end{align}

\textbf{Contradiction Type 2: Directional Contradiction}

This type occurs when the same subject claims contradictory spatial positions relative to the same object.

{Vertical Contradictions:}
\begin{align}
 &\quad \langle \text{"building"}, \text{"top"}, \text{"sky"} \rangle \text{ vs. } \langle \text{"sky"}, \text{"bottom"}, \text{"building"} \rangle 
\end{align}

{Horizontal Contradictions:}
\begin{align}
&\quad \langle \text{"sky"}, \text{"right"}, \text{"building"} \rangle \text{ vs. } \langle \text{"sky"}, \text{"left"}, \text{"building"} \rangle
\end{align}

\subsubsection{Valid vs. Invalid Bidirectional Relations}
We distinguishes between contradictory and complementary spatial relationships.

\textbf{Invalid (Contradictory):}
\begin{align}
&\langle \text{"building"}, \text{"right"}, \text{"sky"} \rangle \\
&\langle \text{"sky"}, \text{"left"}, \text{"building"} \rangle \quad \rightarrow \text{Contradiction Resolution}
\end{align}

\textbf{Valid (Complementary):}
\begin{align}
&\langle \text{"sky"}, \text{"bottom"}, \text{"building"} \rangle \\
&\langle \text{"building"}, \text{"top"}, \text{"sky"} \rangle \quad \rightarrow \text{KEPT}
\end{align}

The key difference is that complementary relations use opposite spatial terms (top/bottom), which is logically consistent, while contradictory relations use the same spatial term in both directions.

\subsubsection{Contradiction Resolution}
After detecting contradictions, we collect all invalid (contradictory) pairs and ask the VLM to select the correct spatial relations.
For each contradictory pair, we present both spatial options to the VLM and let it choose the correct relationship based on visual evidence.

{Resolution Prompt Template:}

\begin{lstlisting}
{caption}From the perspective of someone looking at the image, 
is the {object} {preposition1} or {preposition2} the {subject}?
\end{lstlisting}

\textbf{Example Resolution Question:}

For contradictory pair $\langle \text{"building"}, \text{"above"}, \text{"sky"} \rangle$ vs. $\langle \text{"sky"}, \text{"below"}, \text{"building"} \rangle$:
\begin{lstlisting}
Q: <image>
A lighthouse and a house with lights on are on a hill under a cloudy sky. 
From the perspective of someone looking at the image, is the sky above or below the building?
A: above
\end{lstlisting}

\textbf{Answer Processing Logic:}
\begin{enumerate}
    \item \textbf{Clear Choice}: 
    
        If answer contains only the first option $\rightarrow$ Return first triplet
    
        If answer contains only the second option $\rightarrow$ Return second triplet
    \item \textbf{Unclear}: If answer contains neither option clearly $\rightarrow$ Return empty (discard both)
\end{enumerate}

This resolution step ensures that contradictory spatial relationships are resolved through direct visual verification, maintaining the quality of the final spatial knowledge base.

Finally, the Valid (Complementary) triplets and Invalid (Contradictory) triplets selected after Contradiction Resolution are kept, 
result in
 43 triplets out of 52 triplets.

This process ensures comprehensive spatial relations extraction while maintaining quality through validation and contradiction detection. The structured questioning methodology provides more reliable results than free-form spatial relations generation.


\end{document}